\documentclass[lettersize,journal]{IEEEtran}
\usepackage{amsmath,amsfonts}
\usepackage{algorithmic}
\usepackage{algorithm}
\usepackage{array}
\usepackage[caption=false,font=normalsize,labelfont=sf,textfont=sf]{subfig}
\usepackage{textcomp}
\usepackage{stfloats}
\usepackage{url}
\usepackage{verbatim}
\usepackage{graphicx}
\usepackage{cite}
\usepackage{soul}
\usepackage{booktabs}
\usepackage{multirow}
\usepackage[colorlinks=true, citecolor=blue]{hyperref}
\usepackage{caption}
\usepackage{balance}
\usepackage{cite}
\usepackage{soul}
\usepackage[font=footnotesize,skip=-0pt]{caption}

\begin{document}

\setlength{\abovedisplayskip}{4pt}
\setlength{\belowdisplayskip}{4pt}
\setlength{\abovedisplayshortskip}{2pt}
\setlength{\belowdisplayshortskip}{2pt}
\setlength{\textfloatsep}{10pt}

\title{A Language-Agnostic Hierarchical LoRA-MoE Architecture for CTC-based Multilingual ASR}

\author{Yuang Zheng, Dongxu Chen, Yuxiang Mei, Dongxing Xu, Jie Chen and Yanhua Long.
        % <-this % stops a space
\thanks{This paper was produced by the IEEE Publication Technology Group. They are in Piscataway, NJ.}% <-this % stops a space
\thanks{Manuscript received April 19, 2021; revised August 16, 2021.}}

\markboth{Journal of \LaTeX\ Class Files,~Vol.~14, No.~8, August~2021}%
{Shell \MakeLowercase{\textit{et al.}}: A Sample Article Using IEEEtran.cls for IEEE Journals}

\IEEEpubid{0000--0000/00\$00.00~\copyright~2021 IEEE}

\maketitle

\begin{abstract}
Large-scale multilingual ASR (mASR) models such as Whisper achieve strong performance but incur 
high computational and latency costs, limiting their deployment on resource-constrained edge devices. 
In this study, we propose a lightweight and language-agnostic multilingual ASR system 
based on a CTC architecture with domain adaptation. Specifically, we introduce a Language-agnostic 
Hierarchical LoRA-MoE (HLoRA) framework integrated into an mHuBERT-CTC model, 
enabling end-to-end decoding via LID-posterior-driven LoRA routing. 
The hierarchical design consists of a multilingual shared LoRA for learning language-invariant acoustic 
representations and language-specific LoRA experts for modeling language-dependent characteristics. 
The proposed routing mechanism removes the need for prior language identity information or 
explicit language labels during inference, achieving true language-agnostic decoding. 
Experiments on MSR-86K and the MLC-SLM 2025 Challenge datasets demonstrate that 
HLoRA achieves comparable performance to two-stage inference approaches while reducing RTF by 11.7\% and 8.2\%, respectively, leading to improved decoding efficiency for low-resource mASR applications.
\end{abstract}


\section{Introduction}
\IEEEPARstart{L}{arge}-scale multilingual speech recognition (mASR) models, 
such as Whisper \cite{Whisper} and SeamlessM4T \cite{SeamlessM4T}, have achieved remarkable 
recognition performance by leveraging massive multilingual training data. 
However, the high computational cost and inference latency severely limit their practical deployment, especially on 
resource-constrained edge devices such as hearing aids and mobile platforms. 
This motivates interest in lightweight mASR systems that 
can balance recognition performance with efficiency and real-time requirements.

Most state-of-the-art mASR systems are built upon attention-based encoder--decoder (AED) architectures, with 
Whisper and its variants representing a dominant research direction \cite{OWSM}. 
These systems typically adopt byte-level byte-pair encoding (BBPE) \cite{BBPE} to unify 
token representations across languages and rely on large-scale multilingual data 
and extensive training resources. However, despite their impressive in-domain results, 
such models often suffer significant performance degradation under domain mismatch, 
including variations in speaking style, acoustic conditions, or accents that differ 
from the training data \cite{SHNU,Frisian,EdAcc,LibriSpeech}. 
% For example, Whisper reports a word error rate (WER) 
% of 2.7\% on the LibriSpeech test-clean set \cite{LibriSpeech}, while the WER increases to 19.7\% on the EdAcc 
% accented English corpus \cite{EdAcc}. 
% Similar degradation has also been observed in the MLC-SLM 2025
% Challenge \cite{SHNU}, where Whisper-large-v3 exhibits WER exceeding 20\% for some languages 
% under conversational speech conditions.

To alleviate domain mismatch in multilingual ASR, recent studies have increasingly 
adopted parameter-efficient fine-tuning strategies, with low-rank adaptation (LoRA) \cite{F., T., G.}. Representative approaches \cite{LoRA-Whisper, two-stage, Q.Zhao, SAML} adapt pretrained models 
to specific languages or speaking styles while largely preserving their original 
multilingual capabilities. Although these methods  reduce adaptation costs, 
most existing approaches focus on monolingual \cite{Jia, Mono} or language-dependent LoRA fine-tuning 
\cite{Yer, Rapha,X, B.} and largely overlook the acoustic commonalities shared across languages.
Moreover, many of these methods require explicit language identity (LID) during inference \cite{Sony,A,W}, 
which substantially restricts usability in real-world scenarios. In practical applications, 
users generally prefer language-agnostic mASR systems that do not require manual language 
selection between conversational turns.

Existing language-agnostic mASR approaches generally follow two paradigms. 
The first relies on universal modeling with shared vocabularies, such as unified BBPE 
token in Whisper \cite{Whisper}. While effective at large scale, such approaches 
often struggle to capture fine-grained language-specific acoustic characteristics. 
The second paradigm adopts two-stage inference pipelines, where an LID module first predicts 
the language and then activates a corresponding expert, such as 
router-based mixture-of-experts (MoE) systems \cite{BLR-MoE, LR-MoE, GoogleMoE}. Although effective, these 
pipelines introduce additional inference latency and are prone to error propagation from 
LID to ASR, making them less suitable for low-latency and on-device applications \cite{Rupak, Hongli, Pu}. 
These limitations motivate a lightweight, end-to-end architecture for 
achieving language-agnostic behavior without sacrificing recognition performance or decoding efficiency.

Motivated by these challenges, this work focuses on CTC-based multilingual ASR and adopts 
mHuBERT-CTC as the backbone to develop a lightweight and truly language-agnostic framework 
with non-autoregressive, low-latency decoding.
Our main contributions are summarized as follows:
\begin{itemize}
    \item A \emph{Language-Agnostic Hierarchical LoRA-MoE} architecture is proposed and 
    integrated into an mHuBERT-CTC framework, enabling multilingual ASR with 
    single-pass, end-to-end decoding.
    \item A hierarchical LoRA design combining a \emph{multilingual shared LoRA} and \emph{language-specific 
    LoRA modules} is developed to jointly model language-invariant acoustic features and 
    language-dependent characteristics, achieving improved efficiency and competitive accuracy.
    \item An \emph{LID-posterior-driven LoRA routing mechanism} is introduced, which dynamically 
    activates language-specific LoRA experts from intermediate outputs and removes 
    the need for prior LID information during inference.
    \item A  \emph{CTC-based language-agnostic baseline} is built by adapting mHuBERT-CTC with 
    language-specific LoRA modules and incorporating a two-stage inference 
    strategy for systematic comparison.
\end{itemize}

\begin{figure}[t]
\captionsetup{belowskip=-10pt}
\centerline{\includegraphics[width=0.8\columnwidth,height=4cm]{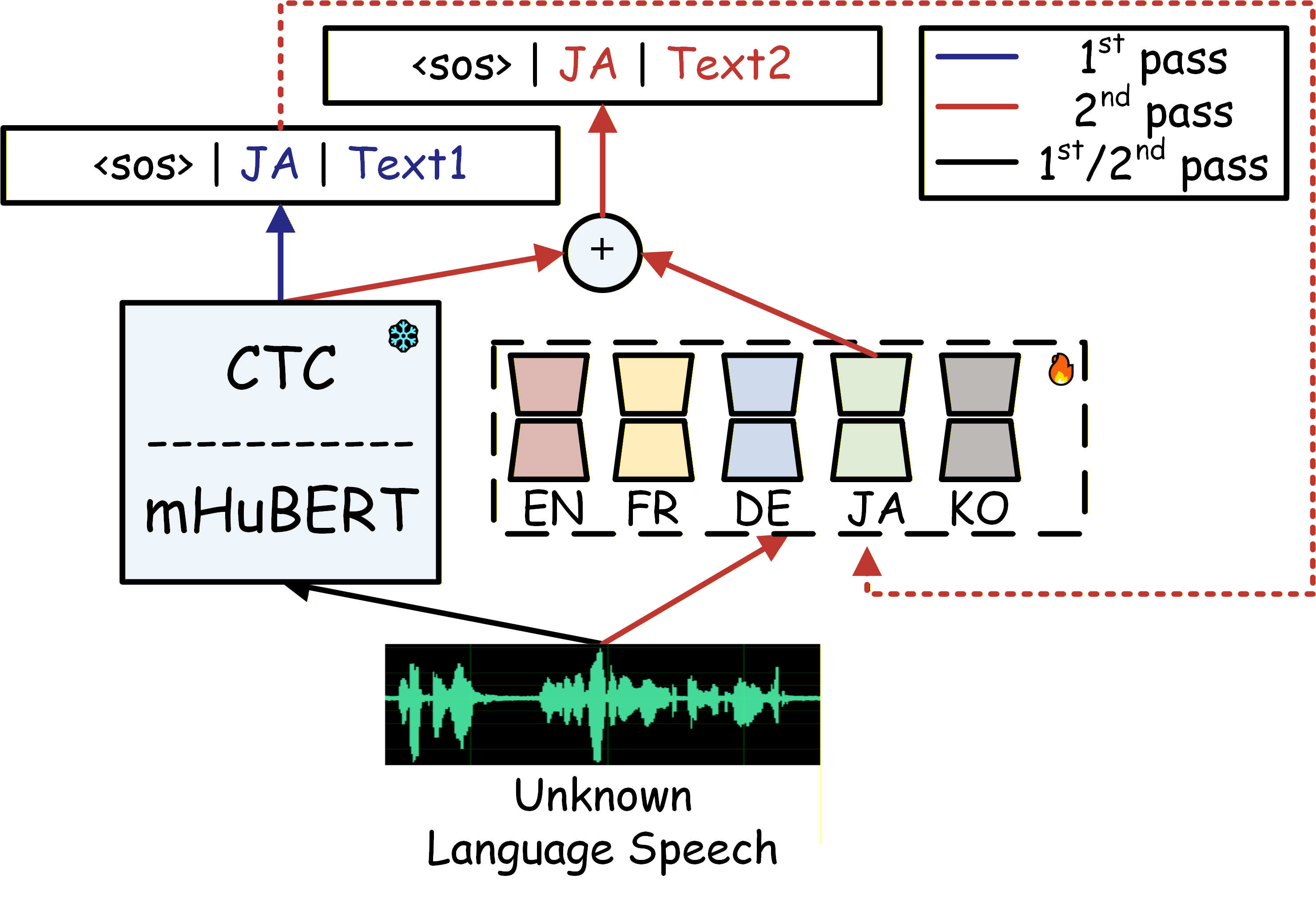}}
\caption{The structure of LoRA-adapted mHubert-CTC mASR with two-stage inference (mHuBERT-CTC-LIDLoRA).}
\label{fig:lidlora}
\end{figure}

\section{LoRA-Adapted mHubert-CTC mASR with Two-Stage Inference}
\label{sec:twostagelora}

To establish a lightweight baseline for language-agnostic multilingual ASR, 
we first build a CTC-based system upon the mHuBERT-147 multilingual pretrained model \cite{mHuBERT}. 
With this backbone, language-specific LoRA modules are introduced for parameter-efficient 
adaptation, and a state-of-the-art two-stage inference strategy \cite{two-stage,LoRA-Whisper} is adopted to 
enable language-agnostic decoding. The resulting system is referred to 
as \emph{mHuBERT-CTC-LIDLoRA}.  The overall architecture is illustrated in Fig.~\ref{fig:lidlora}, and its components are described below.

\textbf{mHuBERT-CTC backbone}: 
To enable implicit language identification within a unified CTC framework, a dedicated 
language token is introduced as the first predict token in the CTC target sequence during 
training. This design allows the model to jointly learn LID and ASR in an E2E manner 
without an explicit LID module.  
The training procedure follows a standard fine-tuning strategy for SSL-pretrained models: 
first optimize the CTC layer while keeping the mHuBERT encoder frozen, and then 
fine-tune the entire network end-to-end.

\textbf{LoRA-adapted mHuBERT-CTC}: 
To improve multilingual adaptability, we adopt 
a LoRA-based fine-tuning strategy. Specifically, 
as shown in Fig.~\ref{fig:lidlora}, independent LoRA modules are trained for 
each target language by freezing all pretrained mHuBERT-CTC parameters and updating only the 
language-specific LoRAs. This results in a  multilingual ASR system 
with explicit language-dependent adaptation capability, while incurring a small number 
of additional trainable parameters.

\textbf{Two-stage language agnostic inference}: 
Language agnostic decoding is achieved through a two-stage inference procedure. 
In the first stage, the base mHuBERT-CTC model without LoRA adaptation is used 
to predict an LID token via CTC decoding, and the language with the highest posterior 
probability is selected. In the second stage, the corresponding language-specific 
LoRA module is activated to perform ASR decoding. Although this two-stage 
approach enables language-agnostic decoding without prior LID information, 
it introduces additional inference latency and is susceptible to error propagation 
from language prediction to ASR. These limitations motivate a more 
efficient single-pass and end-to-end solution.

\section{mHuBERT-CTC with Hierarchical LoRA-MoE Architecture}
\label{sec:loramoe}

While language-specific LoRA adaptation enables parameter efficient multilingual ASR, 
independent per-language training is inefficient and fails to leverage cross-lingual 
acoustic commonalities. Existing two-stage language-agnostic inference further incurs 
redundant computation and increased latency, while relying on frozen LID predictions 
that cannot be improved during downstream adaptation. To address these issues, we propose 
a hierarchical LoRA-MoE architecture built upon mHuBERT-CTC 
(\emph{mHuBERT-CTC-HLoRA}), which jointly optimizes LID and ASR, enables unified modeling of 
language-invariant and language-specific representations, and supports efficient 
single-pass, language-agnostic decoding.

\begin{figure*}[t]
\captionsetup{belowskip=-10pt}
\centerline{\includegraphics[width=1.8\columnwidth,height=4cm]{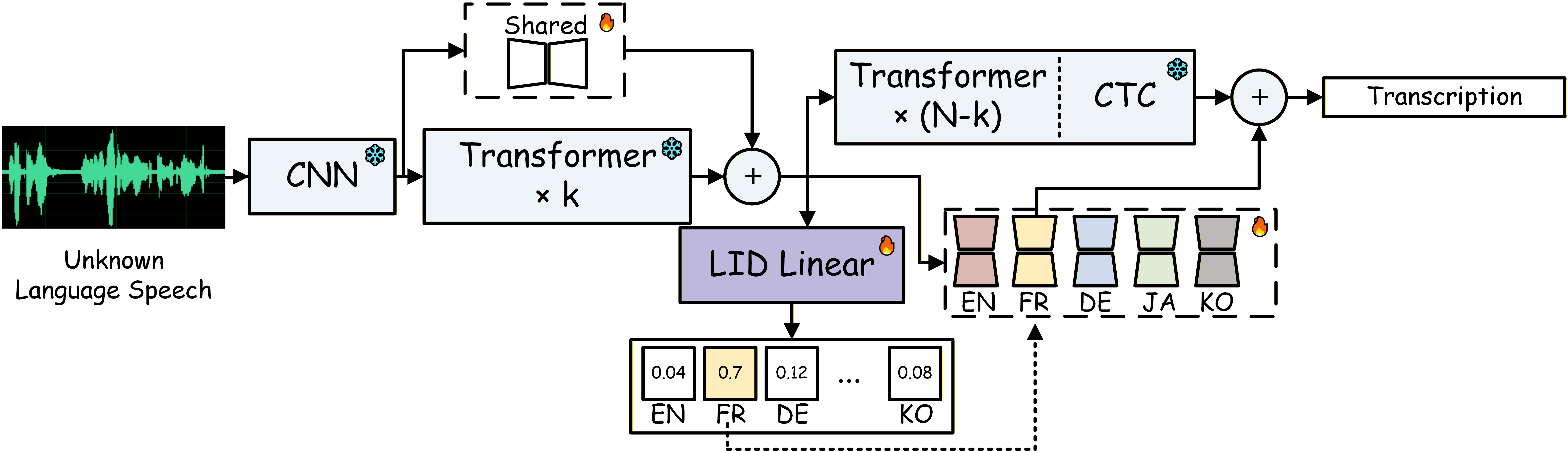}}
\caption{The proposed mHuBERT-CTC-HLoRA architecture. The CNN and Transformer backbone are frozen, while the shared LoRA in the first $k$ Transformer layers, the language-specific LoRA in the last $N\!-\!k$ layers and CTC head, as well as the LID linear layer, are trainable.}
\label{fig:hlora}
\end{figure*}

\subsection{Model Architecture}
\label{subsec:march}

Fig.~\ref{fig:hlora} illustrates the proposed mHuBERT-CTC-HLoRA architecture.
An unknown-language speech is processed by a  frozen CNN front-end
and an $N$-layer Transformer encoder from the pretrained mHuBERT-147 model.
To balance cross-lingual sharing and language-specific modeling,
the encoder is divided into two parts:
the lower $k$ layers incorporate a \emph{shared LoRA} optimized across all languages
to capture language-invariant acoustic representations,
while the upper $N-k$ layers employ \emph{language-specific LoRA} modules
to model discriminative language-dependent characteristics. 
Intermediate representation after $k$-th layer is fed into a LID classifier, whose posterior dynamically activates the corresponding
language-specific LoRA in upper layers and CTC head,
thereby tightly coupling LID and ASR within a unified framework
and enabling single-pass, language-agnostic decoding.

\subsection{Model Training} 
\label{subsec:mtrain}
Let the multilingual training set be denoted as
$D=\{D_{1},D_{2},\dots,D_{L}\}$, where $D_{i}$ contains utterances from
the $i$-th language.
Each language-specific subset is first split into minibatches.
All minibatches across languages are then pooled together and randomly
shuffled at the beginning of every training epoch.
At each iteration, a minibatch from language $i$ is randomly sampled.
The input speech $X$ is processed by the frozen CNN front-end and the
first $k$ Transformer layers augmented with the shared LoRA:
\begin{equation}
    X_{h} = \mathcal{F}_{1}\left(\texttt{CNN} \left( X \right),W_{s},\theta_{trans}^{k}\right) 
\end{equation}
where $W_s$ denotes the shared LoRA parameters and
$\theta^{k}_{\text{trans}}$ are the frozen pretrained weights.
Next, $X_{h}$ is forwarded to the remaining $N-k$ Transformer layers 
with the \emph{language-specific LoRA} $W_{L}^{i}$, selected according to the ground-truth language label: 
\begin{equation}
    X_{N} = \mathcal{F}_{2}\left(X_{h}, W_{L}^{i}, \theta_{trans}^{N-k}\right)
\end{equation}
followed by CTC decoding to compute the ASR loss:
\begin{equation}
    \mathcal{L}_{ASR} = \texttt{CTC}\left(X_{N}, W_{L}^{i}\right)
\end{equation}
In parallel, $X_{h}$ is fed into the LID classifier(a projection layer) to predict the language:
\begin{align}
    Y_{L} & = \texttt{LIDLinear}\left(X_{h}\right) \\
    \mathcal{L}_{LID} & = \texttt{CE} \left(Y_{L},  Y_{L}^{true} \right)
\end{align}
The total training objective is defined as a weighted combination of
ASR and LID losses with $\lambda$:
\begin{align}
    \mathcal{L} = (1-\lambda)\mathcal{L}_{ASR}+\lambda \mathcal{L}_{LID}
\end{align}

Although only one language-specific LoRA $W_L^{i}$ is activated per
iteration, the shared LoRA $W_s$ and the LID classifier are updated in every iteration.
By alternating languages through shuffled multilingual minibatches,
the proposed training strategy enables balanced optimization of all
language-specific LoRA modules while jointly learning shared acoustic
representations and language discrimination.

\subsection{Inference strategy}
\label{subsec:infer}

During inference, the proposed mHuBERT-CTC-HLoRA framework performs
\emph{single-pass language-agnostic decoding} via an
\emph{LID-posterior-driven LoRA routing mechanism}.
Given an input utterance, the shared LoRA modules in the lower Transformer
layers are first applied to extract language-invariant representations.
Based on the intermediate outputs, an integrated LID classifier predicts
a posterior distribution over the target languages.
This LID posterior is then directly used for hard argmax routing to activate the
corresponding language-specific LoRA expert in the upper Transformer layers
and the CTC head within the same forward pass, producing the final transcription.

Unlike the two-stage inference baseline described in
Section~\ref{sec:twostagelora}, which relies on a separate encoder pass and
external LID decision, the proposed approach jointly performs language
identification and ASR in a unified, end-to-end manner.
By leveraging LID posteriors estimated from intermediate shared representations,
HLoRA eliminates the need for prior LID information, alleviates error propagation
across stages, and significantly reduces inference latency.
As a result, the proposed inference strategy enables efficient, low-latency,
and truly language-agnostic multilingual ASR.

%\vspace{-0.2cm}

\section{Experiments and Results}
\subsection{Datasets}

Experiments are conducted on the MSR-86K \cite{MSR-86K} and the MLC-SLM 2025 
Challenge \cite{SHNU} datasets. MSR-86K is used to build the mHuBERT-CTC 
pretrained ASR model and taken as source-domain mASR task. From its 1500-hour subset, 11 languages are selected: 
English (EN), French (FR), German (DE),
Italian (IT), Japanese (JA), Korean (KO), Portuguese (PT), Russian (RU), Spanish (ES),
Thai (TH), and Vietnamese (VI).
MLC-SLM dataset is taken as the low-resource target domain mASR dataset for system evaluation, focusing on 
Indian-accented English (EN-IN), FR, DE, JA, and KO.
For each language, 100 hours of audio are used for training (98h) and development (2h),
while the official development and test sets provided by the challenge 
are used for evaluation.

\subsection{Configurations}

We adopt the pretrained SSL model mHuBERT-147 \cite{mHuBERT} as the backbone.
SentencePiece tokenization uses character units for JA, KO, and TH and a 4k BPE vocabulary
for other languages, resulting in a unified vocabulary of 9,521 tokens.
LoRA is applied to the self-attention Q, K, V projections and the CTC layer
with rank $r=32$ and $\alpha=64$.
Evaluation is performed using mixture error rate, with character error rate (CER) 
for JA/KO/TH and WER otherwise).

\begin{table*}[t]
\renewcommand\arraystretch{0.8}
\caption{Overall mASR results on MLC-SLM datasets.
MSR-1500h and MLC-500h denote training data subsets selected from MSR-86K and MLC-SLM, respectively.
“$\surd$” and “$\times$” indicate language-known and language-agnostic inference,
while “single” and “double” denote single-pass and two-stage inference, respectively.
``RTF'' indicates real-time factor measured on a single NVIDIA RTX 4090 GPU(24G) with beam size 10 and batch size 1.
For S1–S6, all systems share the same mHuBERT-147 backbone pretrained on 90k hours of multilingual data. }
\centering
\begin{tabular}{l|lcccccccc}
\toprule
\multirow{2}*{ID} & \multirow{2}*{System } & \multirow{2}*{Params(M)} & \multirow{2}*{TrainingData} & \multirow{2}*{LID} & \multirow{2}*{Inference} & \multicolumn{2}{c}{RTF} & \multicolumn{2}{c}{Evaluation Sets(WER\%)} \\
\cmidrule(r){7-8} \cmidrule(l){9-10}
~ & ~ & ~ & ~ & ~ & ~ & MLC-dev & MLC-test & MLC-dev & MLC-test \\
\midrule
S1 & \multirow{2}*{mHuBERT-CTC(baseline)} & \multirow{2}*{97} & MSR-1500h & × & single & 0.553 & 0.515 & 44.5 & 43.7 \\
S2 & ~ & ~ & MLC-500h & × & single & 0.549 & 0.514 & 22.5 & 21.6 \\
\midrule
S3 & \multirow{2}*{mHuBERT-CTC-LIDLoRA} & \multirow{2}*{107} & \multirow{2}*{MLC-500h} & $\surd$ & single & 0.563 & 0.539 & 24.6 & 23.0 \\
S4 & ~ & ~ & ~ & × & double & 0.592 & 0.564 & \textbf{26.6} & \textbf{24.8} \\
\midrule
S5 & \multirow{2}*{mHuBERT-CTC-HLoRA} & \multirow{2}*{102} &  \multirow{2}*{MLC-500h} & $\surd$ & single & 0.526 & 0.531 & 26.0 & 24.0 \\
S6 & ~ & ~ & ~ & × & single & 0.523 & 0.518 & \textbf{26.3} & \textbf{24.7} \\
\midrule
S7 & Whisper-base \cite{Whisper} &  74  & 680,000h & × & single & - & - & 37.2 & 35.4 \\
S8 & Whisper-small \cite{Whisper} &  244 & 680,000h & × & single & - & - & 29.7 & 28.6 \\
\bottomrule
\end{tabular}
\label{tab:overall}
\vspace{-16pt}
\end{table*}

\subsection{Overall Results}
\label{subsec:reslt}

Table~\ref{tab:overall} reports overall mASR performance on the MLC-SLM 2025
development and test sets.
S1 is trained on source-domain MSR-1500h, while S2 initializes from S1
and is fully fine-tuned on target-domain MLC-500h.
The large gap between S1 and S2 highlights the severe impact of
domain mismatch on conversational speech.
Moreover, 
state-of-the-art Whisper models (S7 and S8) still underperform S2 on the test set, emphasizing the need for efficient domain adaptation in pretrained mASR systems.

S3 and S4 evaluate LoRA-based parameter-efficient adaptation on the source-domain mHuBERT-CTC baseline (S1), updating only a small subset of parameters.
As expected, this parameter-efficient setting leads to a noticeable performance
gap compared with full fine-tuning in S2.
However, this behavior is reasonable  in low-resource adaptation
scenarios, where full fine-tuning of large pretrained models is computationally
expensive and prone to overfitting to the target domain, potentially degrading
performance on the source domain.

S3 uses language-dependent inference, activating the language-specific LoRA module with the ground-truth label, while S4 performs language-agnostic decoding via a two-stage pipeline.
Although S4 achieves fully language-agnostic operation, its recognition
performance is consistently lower than that of S3.
This gap primarily arises from the two-stage design, where the activation of
language-specific LoRA modules relies on the LID prediction produced by the
base mHuBERT-CTC model. Since this LID component is inherited from a frozen pretrained model and
cannot be jointly optimized during LoRA adaptation, errors in language
prediction directly affect subsequent ASR decoding.
These observations reveal a fundamental limitation of two-stage
language-agnostic inference and motivate the proposed single-pass HLoRA
framework, where LID prediction and ASR are jointly optimized end-to-end.

S5 and S6 evaluate HLoRA under language-known and language-agnostic inference, respectively. S5 shows slightly degraded performance than S3 due to reduced language-specific LoRA capacity. By design, HLoRA allocates fewer language-specific parameters in order to
enable a lightweight and streaming-friendly architecture, trading a small
amount of accuracy for improved efficiency and unified modeling.

More importantly, S6  demonstrates superior efficiency by 
achieving competitive performance via unified single-pass inference. S6 achieves competitive WERs, which are comparable to or slightly better than the two-stage baseline S4, while reducing relative RTF by 11.7\% and 8.2\%. The critical advantage lies in the HLoRA framework’s ability to jointly optimize LID and ASR 
within a single, unified model, thereby enabling single-pass inference.
This design not only 
mitigates the computational overhead and latency associated with two-stage inference but also 
confirms that robust language-aware adaptation can be achieved implicitly, without relying on 
explicit LID input at inference time. These results validate HLoRA as an efficient and practical 
solution for low-latency, language-agnostic multilingual ASR.

\subsection{Ablation Results}

\begin{table}[h]

\renewcommand\arraystretch{0.8}
\caption{mHuBERT-CTC-HLoRA ablation study with different $k$. Language-wise results are on the MLC-dev/MLC-test in WER\%.}
\centering
\begin{tabular}{c|ccccc}
\toprule
 Language & $k = 1$ & 3 & 6 & 9 & 11 \\
\midrule
English & 40.5/48.1 & 33.8/36.6 & 29.0/26.6 & 29.2/\textbf{25.2} & \textbf{28.9}/25.3 \\
French & 26.6/33.9 & 26.3/33.2 & \textbf{26.1}/\textbf{31.7} & 26.4/32.0 & 26.3/32.2 \\
German & 61.2/54.3 & 54.9/44.1 & \textbf{39.1}/32.9 & 39.5/\textbf{32.6} & 39.8/33.4 \\
Japanese & 37.3/41.0 & 32.6/37.0 & \textbf{25.0}/27.8 & 25.5/\textbf{23.6} & 25.7/24.7 \\
Korean & 17.9/16.5 & 17.5/\textbf{16.0} & \textbf{17.2}/16.1 & 17.7/16.6 & 18.1/16.9 \\
\midrule
avg & 34.8/36.6 & 31.3/32.1 & \textbf{26.0}/26.0 & 26.3/\textbf{24.7} & 26.5/25.2 \\
\bottomrule
\end{tabular}
\label{tab:ablation}

\end{table}

Table~\ref{tab:ablation} presents the ablation results of the proposed
mHuBERT-CTC-HLoRA model with different numbers of shared Transformer layers $k$.
The results reveal a clear trade-off between cross-lingual sharing and
language-specific adaptation.
When $k$ is small (e.g., $k=1$ or $k=3$), most Transformer layers rely on
language-specific LoRA modules, leading to insufficient cross-lingual sharing
and inferior performance across all languages.
As $k$ increases, the model benefits from stronger language-invariant
representations learned by the shared LoRA, resulting in substantial WER
reductions.
The best overall performance is achieved with a moderate number of shared
layers ($k=6$ or $k=9$), which yields the lowest average WER on both
MLC-dev and MLC-test. 
Further increasing $k$ to 11 slightly degrades performance, suggesting that
excessive parameter sharing limits the model’s ability to capture
language-specific acoustic characteristics.
These results confirm that an appropriate hierarchical decomposition of shared
and language-specific LoRA modules is critical for balancing model capacity, and efficiency.

\begin{figure}[h]
\vspace{-12pt}
\captionsetup{aboveskip=-8pt, belowskip=-8pt}
\centerline{\includegraphics[width=0.8\columnwidth]{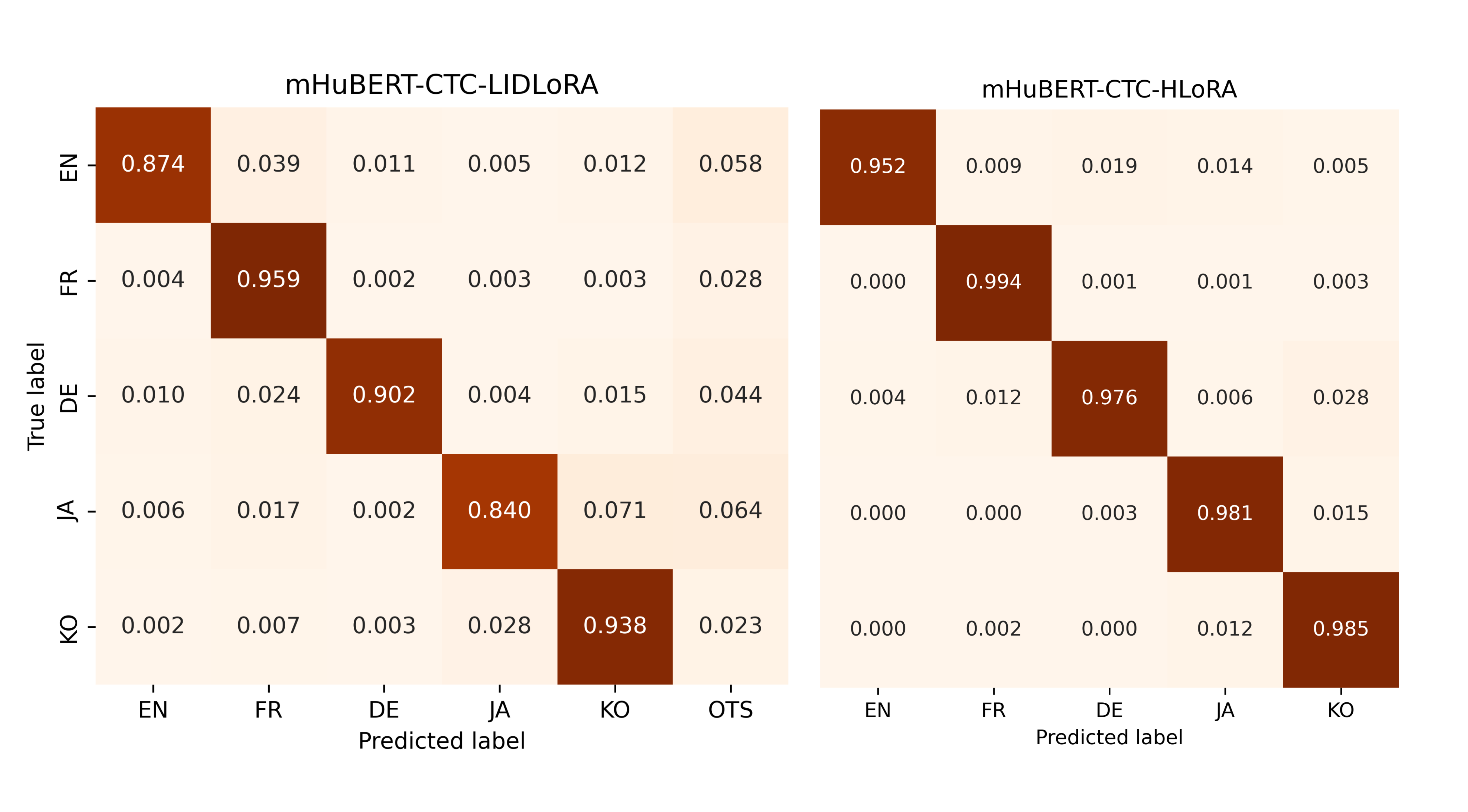}}
\caption{Confusion matrix of LID results on the MLC dev-set
for LIDLoRA (left) and the proposed HLoRA (right).
``OTS'' denotes the aggregated predictions of the remaining six
source-domain languages in the pretrained mHuBERT-CTC model that are not 
included in five target-domain languages.}
\label{fig:confusion}
\end{figure}

In addition, Fig.~\ref{fig:confusion} further compares the LID behavior of LIDLoRA and the proposed HLoRA.
HLoRA produces more diagonal-dominant confusion matrices across all five target 
languages, leading to an average LID accuracy improvement from 90.1\% to 97.9\%.
This gain is attributed to the single-pass, end-to-end HLoRA design, which 
jointly optimizes LID and ASR using shared LoRA features, avoiding the frozen 
and non-adaptive LID stage in two-stage inference and enabling more reliable 
language-agnostic decoding.

\section{Conclusion}
This paper presented HLoRA, a lightweight and streaming-friendly 
multilingual ASR framework designed for low-resource, language-agnostic scenarios. 
By hierarchically decomposing LoRA modules into shared and language-specific 
components, the proposed method effectively balances cross-lingual 
representation sharing and language-dependent acoustic modeling. Extensive 
experiments on the MLC-SLM 2025 benchmark demonstrate that HLoRA consistently 
outperforms strong two-stage language-agnostic baselines while avoiding external 
LID supervision. Moreover, the single-pass design significantly improves LID 
robustness and inference efficiency, making the system more suitable for 
practical multilingual deployment. Our models and source code are publicly released at https://github.com/zhengyuang7/HLoRA
to facilitate reproducibility and future research.


\begin{thebibliography}{32}
\bibliographystyle{IEEEtran}

\bibitem{Whisper}A. Radford, J. W. Kim, T. Xu, G. Brockman, C. Mcleavey, and I. Sutskever, ``Robust Speech Recognition via Large-Scale Weak Supervision'' in \emph{Proc. of the International Conf. on Machine Learning (ICML),} 2023, pp. 28492-28518.

\bibitem{SeamlessM4T}SeamlessM4T, Seamless is a family of AI models that enable more natural and authentic communication across languages. [Online]. Available:
\url{https://github.com/facebookresearch/seamless_communication}

\bibitem{OWSM}Y. Peng, S. Muhammad, Y. Sudo, W. Chen, J. Tian, C.-J. Lin, and S. Watanabe, ``OWSM v4: Improving Open Whisper-Style Speech Models via Data Scaling and Cleaning,'' in \emph{Proc. of the Conf. of the International Speech Communication Association (INTERSPEECH),} 2025.

\bibitem{BBPE}R. Sennrich, B. Haddow, and A. Birch, ``Neural Machine Translation of Rare Words with Subword Units,'' in \emph{Proc. of the 54th Annual Meeting of the Association for Computational Linguistics (ACL),} Vol. 1, pp. 1715–1725.

\bibitem{SHNU}Y. Mei, Y. Zheng, D. Xu, and Y. Long, ``SHNU Multilingual Conversational Speech Recognition System for INTERSPEECH 2025 MLC-SLM Challenge,'' in \emph{Proc. of Workshop on Multilingual Conversational Speech Language Model (MLC-SLM),} 2025, pp. 38-42. 

\bibitem{Frisian}R. Amooie, W. D. Vries, Y. Hao, J. Dijkstra, M. Coler, and M. Wieling, ``Enhancing Standard and Dialectal Frisian ASR: Multilingual Fine-tuning and Language Identification for Improved Low-resource Performance'' in \emph{Proc. of IEEE International Conf. on Acoustics, Speech, and Signal Processing (ICASSP),} 2025. 

\bibitem{EdAcc}R. Sanabria, N. Bogoychev, N. Markl, A. Carmantini, O. Klejch, and P. Bell, ``The Edinburgh International Accents of English Corpus: Towards the Democratization of English ASR,'' in \emph{Proc. of IEEE International Conf. on Acoustics, Speech, and Signal Processing (ICASSP),} 2023, pp. 1-5.

\bibitem{LibriSpeech}V. Panayotov, G. Chen, D. Povey, and S. Khudanpur, ``Librispeech: An ASR corpus based on public domain audio books,'' in \emph{Proc. of IEEE International Conf. on Acoustics, Speech, and Signal Processing (ICASSP),} 2015.

\bibitem{F.}F. Zhang, W. Geng, H. Huang, Y. Shan, C. Yi, and H. Qu, ``Boosting Code-Switching ASR with Mixture of Experts Enhanced Speech-Conditioned LLM,'' in \emph{Proc. of IEEE International Conf. on Acoustics, Speech, and Signal Processing (ICASSP),} 2025, pp. 1-5.

\bibitem{T.}T. P. Ferraz, Marcely Zanon Boito, Caroline Brun, and Vassilina Nikoulina, ``Multilingual Distilwhisper: Efficient Distillation of Multi-Task Speech Models Via Language-Specific Experts,'' in \emph{Proc. of IEEE International Conf. on Acoustics, Speech, and Signal Processing (ICASSP),} 2024, pp. 10716-10720.

\bibitem{G.}G. Kim, B. Lee, D. Kim, and H. Ko, ``Gated Low-Rank Adaptation for Personalized Code-Switching Automatic Speech Recognition on the Low-Spec Devices,'' in \emph{Proc. of IEEE International Conf. on Acoustics, Speech, and Signal Processing (ICASSP),} 2024, pp. 760-764.


\bibitem{LoRA-Whisper}Z. Song, J. Zhuo, Y. Yang, Z. Ma, S. Zhang, and X. Chen, ``LoRA-Whisper: Parameter-Efficient and Extensible Multilingual ASR,'' in \emph{Proc. of the Conf. of the International Speech Communication Association (INTERSPEECH),} 2024, pp. 3934-3938. 

\bibitem{two-stage}C. Y. Kwok, H. Liu, J. Q. Yip, S. Li, and E. S. Chng, ``A Two-Stage LoRA Strategy for Expanding Language Capabilities in Multilingual ASR Models,'' in \emph{IEEE Transactions on Audio, Speech and Language Processing (TASLP)}, pp. 1-16, 2025.

\bibitem{Q.Zhao}Q. Zhao, G. Sun, C. Zhang, M. Xu, and F. Zheng, ``Speaker Adaptive Mixture of Weight-Decomposed LoRA Experts for On-Device End-to-End ASR,'' in \emph{IEEE Transactions on Audio, Speech and Language Processing (TASLP),} pp. 2485-2496, 2025.

\bibitem{SAML}Q. Zhao, G. Sun, C. Zhang, M. Xu, and T. F. Zheng, ``SAML: Speaker Adaptive Mixture of LoRA Experts for End-to-End ASR,'' in \emph{Proc. of the Conf. of the International Speech Communication Association (INTERSPEECH),} 2024, pp. 777-781.

% \bibitem{Khe}Khe Chai Sim, Zhouyuan Huo, Tsendsuren Munkhdalai, Nikhil Siddhartha, Adam Stooke, Zhong Meng, Bo Li, and Tara Sainath, ``A Comparison of Parameter-Efficient ASR Domain Adaptation Methods for Universal Speech and Language Models,'' in \emph{Proc. IEEE International Conf. on Acoustics, Speech, and Signal Processing (ICASSP),} 2024, pp. 6900-6904.

\bibitem{Jia}J. Li, Y. Shao, J. Zhuo, C. Li, L. Tang, D. Yu, and Y. Qian, ``Efficient Multilingual ASR Finetuning via LoRA Language Experts,''  in \emph{Proc. of the Conf. of the International Speech Communication Association (INTERSPEECH),} 2025, pp. 1138-1142.


\bibitem{Mono}O. Babatunde, V. Olufemi, E. Bolarinwa, K. Moshood, and C. Emezue, ``Beyond Monolingual Limits: Fine-Tuning Monolingual ASR for Yoruba-English Code-Switching,'' in \emph{Proc. of Workshop on Computational Approaches to Linguistic Code-Switching,}, 2025, pp. 18-25.

\bibitem{B.}B. Mu, K. Wei, Q. Shao, Y. Xu, and L. Xie, ``HDMoLE: Mixture of LoRA Experts with Hierarchical Routing and Dynamic Thresholds for Fine-Tuning LLM-based ASR Models,'' in \emph{Proc. of IEEE International Conf. on Acoustics, Speech, and Signal Processing (ICASSP),} 2025, pp. 1-5.






\bibitem{Yer}Y. Khassanov, Z. Chen, T. Chen, T. Y. Chong, W. Li, J. Zhang, L. Lu, and Y. Wang, ``Dual-Pipeline with Low-Rank Adaptation for New Language Integration in Multilingual ASR,'' in \emph{Proc. of the Conf. of the International Speech Communication Association (INTERSPEECH),} 2024.


\bibitem{Rapha}R. Bagat, I. Illina, and E. Vincent, ``Mixture of LoRA Experts for Low-Resourced Multi-Accent Automatic Speech Recognition,'' in \emph{Proc. of the Conf. of the International Speech Communication Association (INTERSPEECH),} 2025.



\bibitem{X}X. Zhuang, Y. Qian, S. Xu, and M. Wang, ``Joint Training Framework for Accent and Speech Recognition Based on Conformer Low-Rank Adaptation,'' in \emph{Proc. of IEEE International Conf. on Acoustics, Speech, and Signal Processing (ICASSP),} 2025, pp. 1-5.



\bibitem{Sony}Y. Kashiwagi, H. Futami, E. Tsunoo, S. Arora, and S. Watanabe, ``Rapid Language Adaptation for Multilingual E2E Speech Recognition Using Encoder Prompting'' in \emph{Proc. of the Conf. of the International Speech Communication Association (INTERSPEECH),} 2024.


\bibitem{A}A. Baby, G. Joseph, and S. Singh, ``Robust Speaker Personalisation Using Generalized Low-Rank Adaptation for Automatic Speech Recognition,'' in \emph{Proc. of IEEE International Conf. on Acoustics, Speech, and Signal Processing (ICASSP),} 2024, pp. 11381-11385.

\bibitem{W}W. Liu, Y. Qin, Z. Peng, and T. Lee, ``Sparsely Shared Lora on Whisper for Child Speech Recognition,'' in \emph{Proc. of IEEE International Conf. on Acoustics, Speech, and Signal Processing (ICASSP),} 2024, pp. 11751-11755.

\bibitem{BLR-MoE}G. Ma, W. Wang, L. Zhou, Y. Yang, Y. Li, and B. Du, ``BLR-MoE: Boosted Language-Routing Mixture of Experts for Domain-Robust Multilingual E2E ASR,'' in \emph{Proc. of IEEE International Conf. on Acoustics, Speech, and Signal Processing (ICASSP),} 2025, pp. 1-5.



\bibitem{LR-MoE}W. Wang, G. Ma, Y. Li, and B. Du, ``Language-Routing Mixture of Experts for Multilingual and Code-Switching Speech Recognition''  in \emph{Proc. of the Conf. of the International Speech Communication Association (INTERSPEECH),} 2023, 10.48550/arXiv.2307.05956.

\bibitem{GoogleMoE}K. Hu, B. Li, T. N. Sainath, Y. Zhang, and F. Beaufays, ``Mixture-of-Expert Conformer for Streaming Multilingual ASR,'' in \emph{Proc. of the Conf. of the International Speech Communication Association (INTERSPEECH),} 2023, pp. 3327-3331. 

% \bibitem{C.}Christopher Simic, Korbinian Riedhammer, and Tobias Bocklet, ``Adapter-Based Multi-Agent AVSR Extension for Pre-Trained ASR Models,'' in \emph{Proc. IEEE International Conf. on Acoustics, Speech, and Signal Processing (ICASSP),} 2025, pp. 1-5.


\bibitem{Rupak}R. R. Ghimire, P. Poudyal, and B. K. Bal, ``Improving on the Limitations of the ASR Model in Low-Resourced Environments Using Parameter-Efficient Fine-Tuning,'' in \emph{Proc. of the 21st International Conf. on Natural Language Processing (NLP),} 2024, pp. 408--415.


\bibitem{Hongli}H. Yang, S. Li, H. Huang, A. Tuohan, and Y. Peng, ``Language-Aware Prompt Tuning for Parameter-Efficient Seamless Language Expansion in Multilingual ASR,'' in \emph{Proc. of the Conf. of the International Speech Communication Association (INTERSPEECH),} 2024. 


\bibitem{Pu}P. Wang, S. Watanabe, and H. V. hamme, ``SSVD: Structured SVD for Parameter-Efficient Fine-Tuning and Benchmarking under Domain Shift in ASR,'' in \emph{Proc. of IEEE Automatic Speech Recognition and Understanding Workshop (ASRU),} 2025.

% \bibitem{B.Mu}Bingshen Mu, Kun Wei, Pengcheng Guo, and Lei Xie, ``Mixture of LoRA Experts With Multi-Modal and Multi-Granularity LLM Generative Error Correction for Accented Speech Recognition,'' in \emph{Proc. IEEE International Conf. on Acoustics, Speech, and Signal Processing (ICASSP),} 2025, pp. 2973-2985.

% \bibitem{Yu}Yu Yu, Chao-Han Huck Yang, Jari Kolehmainen, Prashanth G. Shivakumar, Yile Gu, Sungho Ryu, Roger Ren, Qi Luo, Aditya Gourav, I-Fan Chen, Yi-Chieh Liu, Tuan Dinh, Ankur Gandhe, Denis Filimonov, Shalini Ghosh, Andreas Stolcke, Ariya Rastow, and Ivan Bulyko, ``Low-rank Adaptation of Large Language Model Rescoring for Parameter-Efficient Speech Recognition,''  in \emph{Proc. IEEE Automatic Speech Recognition and Understanding Workshop (ASRU),} 2023.


% \bibitem{Minh}Minh Tran, Yutong Pang, Debjyoti Paul, Laxmi Pandey, Kevin Jiang, Jinxi Guo, Ke Li, Shun Zhang, Xuedong Zhang, and Xin Lei, ``A Domain Adaptation Framework for Speech Recognition Systems with Only Synthetic data,'' in \emph{Proc. IEEE International Conf. on Acoustics, Speech, and Signal Processing (ICASSP),} 2025, pp. 1-5.


\bibitem{mHuBERT}M. Z. Boito, V. Iyer, N. Lagos, L. Besacier, and I. Calapodescu, ``mHuBERT-147: A Compact Multilingual HuBERT Model,'' in \emph{Proc. of the Conf. of the International Speech Communication Association (INTERSPEECH),} 2024.


\bibitem{MSR-86K}S. Li, Y. You, X. Wang, Z. Tian, K. Ding, and G. Wan, ``MSR-86K: An Evolving, Multilingual Corpus with 86,300 Hours of Transcribed Audio for Speech Recognition Research,'' in \emph{Proc. of the Conf. of the International Speech Communication Association (INTERSPEECH),} 2024. 










\end{thebibliography}
\end{document}